\documentclass[letterpaper, 10 pt, conference]{ieeeconf}  

\IEEEoverridecommandlockouts                              

\title{\LARGE \bf
AgenticSwarm: Semantic Perception and Adaptive Task Allocation for Heterogeneous Multi-UAV Missions
}

\author{Muhammad Ahsan Mustafa, Yasheerah Yaqoot, Faryal Batool, Roohan Ahmed Khan, \\ Valerii Serpiva, and Dzmitry Tsetserukou
}

\usepackage{graphicx}
\usepackage{booktabs}
\usepackage{amssymb}
\usepackage{makecell}
\usepackage{tabularx}
\usepackage{amsmath}
\usepackage{titlesec}
\let\labelindent\relax
\usepackage{enumitem}
\usepackage{caption}
\usepackage{float}
\usepackage[font=small]{caption}
\usepackage{titlesec}
\usepackage{subcaption}
\usepackage[table]{xcolor}
\usepackage{pifont}
\usepackage{wrapfig}
\usepackage{tabularx}
\usepackage[most]{tcolorbox}
\usepackage{xcolor}
\usepackage{algorithm}
\usepackage{algpseudocode}
\usepackage{placeins}
\usepackage{array}
\usepackage{tabularx}
\usepackage{booktabs}
\usepackage{balance}

\newcolumntype{Y}{>{\centering\arraybackslash}X}

\newtcolorbox{promptbox}[1]{
    colback=gray!4,
    colframe=gray!55,
    title=#1,
    fonttitle=\bfseries,
    coltitle=black,
    boxrule=0.4pt,
    arc=1.5mm,
    left=1mm,
    right=1mm,
    top=1mm,
    bottom=1mm,
    before skip=0.6em,
    after skip=0.6em,
    breakable
}

\newcolumntype{C}{>{\centering\arraybackslash}X}

\titlespacing*{\section}
{0pt}{0.5ex plus 0.1ex minus 0.1ex}{0.3ex plus 0.1ex}

\titlespacing*{\subsection}
{0pt}{0.4ex plus 0.1ex minus 0.1ex}{0.2ex plus 0.1ex}

\titlespacing*{\subsubsection}
{0pt}{0.3ex plus 0.1ex minus 0.1ex}{0.15ex plus 0.05ex}

\setlist[itemize]{topsep=2pt, itemsep=1pt, parsep=0pt, leftmargin=*}
\setlist[enumerate]{topsep=2pt, itemsep=1pt, parsep=0pt, leftmargin=*}

\AtBeginDocument{%
  \setlength{\abovedisplayskip}{3pt}
  \setlength{\belowdisplayskip}{3pt}
  \setlength{\abovedisplayshortskip}{1pt}
  \setlength{\belowdisplayshortskip}{1pt}
}

\begin{document}

\maketitle
\thispagestyle{empty}
\pagestyle{empty}

\begin{abstract}

Multi UAV missions in complex environments require the system to understand both the surrounding scene and the intent of a human operator while maintaining feasible task allocation as mission conditions change. This paper presents AgenticSwarm, an agentic framework for semantic perception and adaptive task allocation in heterogeneous multi UAV missions. An agent interprets aerial imagery and natural language instructions to construct a grounded mission representation that links perceived objects and regions with task requirements, capability constraints, and mission dependencies. This information augments a constrained task allocation process in which obstacle aware path feasibility, energy consumption, and protected return home requirements are incorporated before assignment. During execution, changes such as UAV failure, battery degradation, or task modification trigger residual mission reconstruction from the current system state, while completed work and reconnaissance progress are retained. AgenticSwarm is evaluated across five diverse Gazebo environments and an indoor real test environment, demonstrating its ability to connect semantic reasoning with constrained allocation and adaptive multi UAV mission execution. Compared with a
Grounding DINO+SAM~2.1 perception baseline, the SAM3-based pipeline improves
class-aware recall by $25.2$ percentage points (pp) and semantic label accuracy by
$29.5$ pp. Ablating residual mission replanning increases mean
repeated work from $0\%$ to $61.7\%$ and post-event recovery time by $58.6\%$,
highlighting the contribution of adaptive replanning to mission execution.

\end{abstract}



\section{Introduction}

Multi UAV systems are increasingly used in inspection, monitoring, logistics, and emergency response, where several vehicles must coordinate under limited sensing, energy, and mission resources. As these missions become more complex, autonomy requires the system to understand the operating environment, interpret high level human instructions, and assign tasks according to the capabilities and physical state of the available UAVs.

Recent foundation model based approaches have enabled robots to connect visual semantics with natural language \cite{vlmap,citynav}, while large language models (LLMs) have been used for task decomposition and multi robot coordination \cite{smartllm,coherent,roco}. Related UAV frameworks have further explored language driven mission planning and execution \cite{mutpllm,tacos}. In parallel, optimization based methods have addressed heterogeneous task allocation, dynamic replanning, and energy constrained mission execution \cite{lipllm,pathmatrix,dynamictask}. However, these directions are commonly studied separately. Semantic navigation methods primarily determine where a robot should move, while task allocation methods generally assume that mission targets and their spatial properties are already available to the planner.
\begin{figure}[t]
    \centering
    \includegraphics[width=\columnwidth]{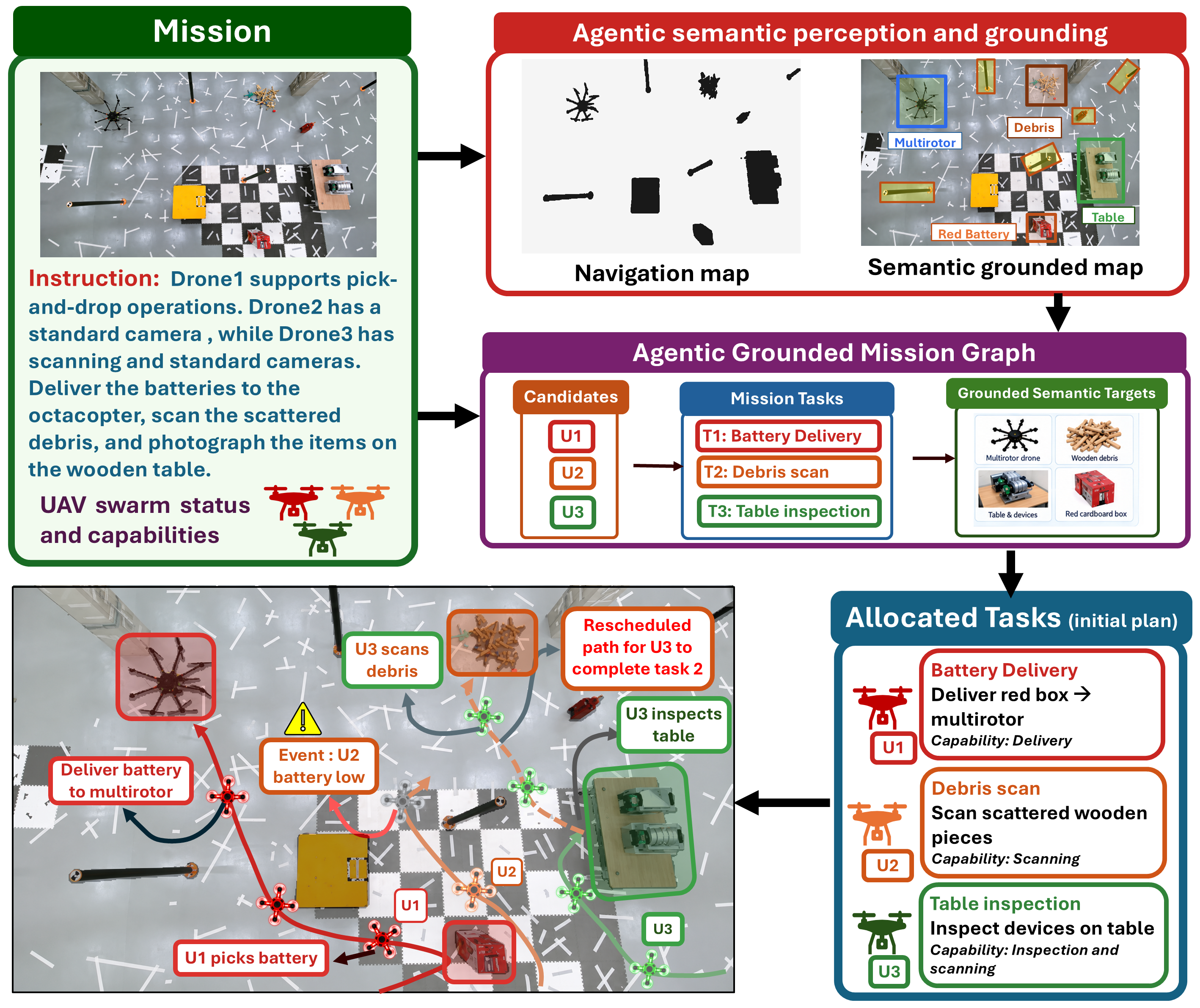}
    \caption{AgenticSwarm overview. A human provides a high level mission, an AI agent performs semantic grounding and task allocation over the perceived environment, and the heterogeneous UAV team executes the mission with residual replanning under runtime changes.}
    \label{fig:main_figure}
    \vspace{-0.2cm}
\end{figure}

This work presents \textit{AgenticSwarm}, an agent based framework that connects semantic environmental perception with natural language mission understanding and constrained task allocation for heterogeneous UAV teams. Visual observations are converted into semantic and navigation representations from which mission relevant objects and regions can be grounded. The resulting mission structure captures task requirements, UAV capabilities, dependencies, and execution state. Obstacle aware route feasibility, estimated energy use, and return requirements are evaluated before constrained optimization determines the task assignments. During execution, changes such as UAV failure, battery variation, or mission updates trigger reconstruction of the remaining mission while previously completed work is retained.
The main contributions of this work are:

\begin{itemize}
    \item An agent based perception and mission reasoning framework that grounds
    natural language UAV missions onto semantic objects and regions identified
    from visual environmental observations.

    \item A constrained heterogeneous task allocation process that incorporates
    capability eligibility, mission dependencies, obstacle aware route
    feasibility, energy requirements, and return feasibility.

    \item A residual mission reconstruction mechanism that preserves completed
    tasks and spatial progress while adapting the remaining mission to changes
    in UAV or task state.

    \item Experimental validation of the complete framework across diverse
    simulated scenarios and an indoor physical test setting, evaluating
    semantic mission grounding, task allocation, and runtime adaptation.
\end{itemize}




\section{Related Work}
\label{sec:literature}

\subsection{Semantic Perception and Language Grounding}
Semantic mapping has increasingly incorporated vision and language models to connect natural language concepts with spatial representations. VLMaps embeds open vocabulary visual language features within a geometric map, enabling semantic landmarks and spatial relations to be queried through language \cite{vlmap}. More recently, CityNavAgent applies open vocabulary perception to aerial navigation by combining scene understanding, object grounding, semantic mapping, and hierarchical language reasoning over urban environments \cite{citynav}. These works demonstrate the importance of spatially grounding semantic information for robotic navigation. Their primary objective, however, is to determine navigation goals for an individual robot, while the perceived entities themselves are not used to construct heterogeneous multi robot mission allocation problems.

\subsection{Language Guided Multi Robot Planning}
LLMs have also been explored for interpreting high level objectives and coordinating robot teams. SMART LLM performs task decomposition and capability aware allocation through language model reasoning \cite{smartllm}, while COHERENT introduces centralized task assignment with execution feedback from heterogeneous robots \cite{coherent}. RoCo uses dialogue between robot specific language agents to negotiate collaborative actions, followed by explicit feasibility validation and conventional motion planning \cite{roco}. The influence of centralized, decentralized, and hybrid language model architectures on coordination performance and scalability has further been examined in \cite{scalable}.

Similar ideas have recently been extended to UAV systems. MUTP LLM combines natural language interpretation with hierarchical task planning and waypoint grounding \cite{mutpllm}, while TACOS separates mission coordination from execution supervision and supports adaptation to changes in vehicle availability \cite{tacos}. SLM A* uses a compact language model to propose multi UAV waypoint sequences that undergo deterministic geometric validation before execution \cite{slmastar}. Collectively, these studies show that language models can provide flexible semantic reasoning for robot teams, although the extent of their authority over task assignment and spatial planning varies considerably.

\subsection{Optimization Based Task Allocation and Adaptation}
Several approaches combine language reasoning with explicit optimization. LiP LLM converts natural language missions into skills and dependency graphs before performing optimization based allocation \cite{lipllm}. Digital twin based planning similarly uses language models to interpret mission updates while integer programming determines the resulting allocation and schedule \cite{digitaltwins}. Hierarchical language model architectures have also been coupled with numerical optimization for adaptive multi robot target tracking \cite{hierarchicalllms}. Other work uses environmental specifications to trigger language guided task inference followed by formal verification and allocation \cite{llmguided}, while \cite{milp} investigates direct generation of mathematical scheduling formulations from natural language. These studies establish that separating semantic reasoning from optimization is already an active direction in multi robot planning.

Dynamic task allocation has also been extensively studied without language models. Path aware approaches couple task assignment with executable route information and preserve mission state during replanning \cite{pathmatrix}. Distributed persistent monitoring methods incorporate energy availability and safe return considerations when redistributing responsibilities \cite{disttask}. Rolling horizon optimization has been used to jointly update task assignments, routes, and charging decisions as task demands evolve \cite{dynamictask}. Energy aware allocation has further been formulated for heterogeneous robots with different capabilities and operating modes \cite{energy}. These methods demonstrate that energy constraints, dynamic reassignment, and path dependent allocation are established components of multi robot mission planning.

AgenticSwarm builds on these complementary research directions by connecting perception derived semantic entities with language based mission interpretation and constrained multi UAV allocation. Semantic objects and regions identified from the environment are grounded into mission tasks, while obstacle aware navigation costs and return feasibility inform the allocation process. During execution, changes in mission or vehicle state lead to reconstruction of the remaining mission while preserving completed task and spatial progress. 




\section{Methodology}
\label{sec:methodology}

\begin{figure*}[t]
    \centering
    \includegraphics[width=0.98\textwidth]{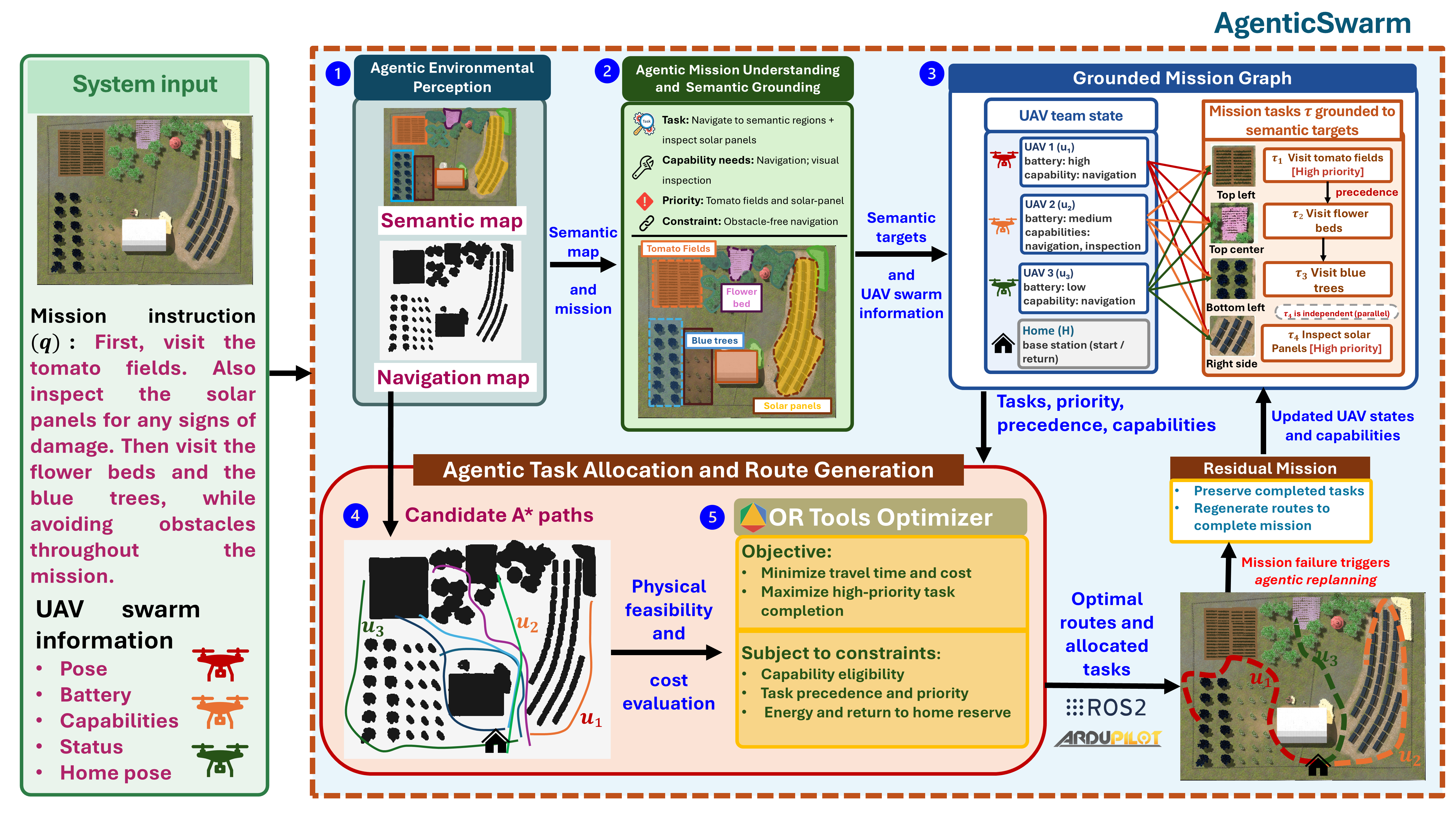}
    \caption{System architecture of AgenticSwarm.}
    \label{fig:system_architecture}
    \vspace{-0.4cm}
\end{figure*}

Fig.~\ref{fig:system_architecture} summarizes the complete AgenticSwarm 
framework. The system receives an aerial RGB observation, a natural language 
mission instruction, and the state of a heterogeneous UAV team. Agentic 
reasoning is used to interpret the observed environment and mission intent, 
while explicit planning and optimization mechanisms determine physical 
feasibility and final task allocation. The complete pipeline consists of 
semantic environmental perception, mission grounding, physically constrained 
allocation, mission execution, and residual replanning.


\subsection{Problem Formulation}
\label{subsec:problem_formulation}
Let the system input be represented as:
\begin{equation}
    \mathcal{X}
    =
    \left\{
        I,\,
        q,\,
        \mathcal{U}
    \right\},
    \label{eq:system_input}
\end{equation} where $I$ denotes the aerial RGB observation, $q$ is the natural language 
mission instruction, and:
\begin{equation}
    \mathcal{U}=\{u_1,u_2,\ldots,u_N\},
\end{equation}
is the available UAV team. Each UAV $u_i$ is associated with a runtime state:
\begin{equation}
    \mathbf{x}_i =
    \left(
        \mathbf{p}_i,\,
        b_i,\,
        \mathcal{C}_i,\,
        s_i
    \right),
    \label{eq:uav_state}
\end{equation}
where $\mathbf{p}_i$ is the current position, $b_i$ is the available battery 
state, $\mathcal{C}_i$ denotes its mission capabilities, and $s_i$ represents 
its operational state.

Given $\mathcal{X}$, AgenticSwarm seeks an executable mission plan that assigns mission tasks to feasible UAVs while satisfying semantic, physical, and mission-level constraints.

\subsection{Semantic Environmental Perception}
\label{subsec:environmental_perception}
The perception stage converts the aerial RGB observation into two complementary 
representations: a semantic representation used for mission understanding and a 
navigation representation used for physical planning.

A vision language model (VLM), in our case GPT-5.6-sol, first analyzes the scene to obtain an open semantic 
inventory of potentially relevant objects and regions. This information guides 
SAM3~\cite{carion2026sam3} based promptable segmentation. The resulting masks subsequently undergo a 
semantic audit in which inconsistent labels, missed entities, and ambiguous 
regions can be reconsidered through targeted reasoning and segmentation. The 
scene profile provides contextual guidance to this process while remaining an 
open prior rather than a fixed detection vocabulary.
The resulting semantic map can be expressed as:
\begin{equation}
    \mathcal{M}_{\mathrm{sem}}
    =
    \left\{
        o_k =
        \left(
            l_k,\,
            m_k,\,
            \mathbf{c}_k
        \right)
    \right\}_{k=1}^{K},
    \label{eq:semantic_map}
\end{equation}
where $l_k$ denotes the semantic label of object or region $o_k$, $m_k$ is its 
pixel mask, and $\mathbf{c}_k$ denotes its spatial reference in image 
coordinates.
A deterministic semantic to occupancy conversion then generates the planning 
map:
\begin{equation}
    \mathcal{M}_{\mathrm{nav}}
    =
    \Phi
    \left(
        \mathcal{M}_{\mathrm{sem}}
    \right),
    \label{eq:semantic_to_nav}
\end{equation}
where $\Phi(\cdot)$ converts semantic regions into free, occupied, or uncertain 
navigation cells according to their planning semantics. Obstacle inflation is 
applied before route planning to provide clearance around occupied regions.

The semantic and navigation representations remain distinct. Semantic object 
identities are retained for mission grounding, while the navigation map 
provides simplified geometry for path planning. If a mission refers to an 
entity that was not retained during generic scene interpretation, targeted SAM3 
grounding can be invoked for that mission entity without automatically 
modifying the trusted navigation representation.

\subsection{Mission Understanding and Semantic Grounding}
\label{subsec:mission_grounding}
The natural language mission $q$ is interpreted in the context of 
$\mathcal{M}_{\mathrm{sem}}$. The reasoning layer extracts the requested task 
semantics and determines the capabilities required for their execution. Mission 
relationships such as precedence, priority, timing preferences, and optional or 
required status can also be associated with the corresponding tasks.

Let the interpreted mission specification be:
\begin{equation}
    \mathcal{Q}
    =
    \left\{
        \tau_j,\,
        \mathcal{C}^{\mathrm{req}}_j,\,
        \rho_j,\,
        \mathcal{D}_j
    \right\}_{j=1}^{M},
    \label{eq:mission_spec}
\end{equation}
where $\tau_j$ denotes task $j$, 
$\mathcal{C}^{\mathrm{req}}_j$ is its required capability set, $\rho_j$ 
represents its mission importance or priority, and $\mathcal{D}_j$ contains 
task relationships such as precedence or other execution constraints.

Semantic references contained in $\tau_j$ are then associated with entities in 
$\mathcal{M}_{\mathrm{sem}}$. An instruction referring to a class of objects 
can therefore expand into multiple spatially distinct executable targets. 
Reconnaissance regions are similarly decomposed into stable spatial patches. 
These patches provide persistent spatial identifiers so that completed coverage 
can remain recorded if the mission is replanned later.

The AI reasoning layer is restricted to semantic interpretation and mission 
structure generation. Final UAV assignment, route feasibility, and physical 
execution decisions are determined by the subsequent planning stages.

\subsection{Grounded Mission Graph Construction}
\label{subsec:mission_graph}
The grounded mission is represented as a typed graph:
\begin{equation}
    \mathcal{G}
    =
    \left(
        \mathcal{V},
        \mathcal{E}
    \right),
    \label{eq:mission_graph}
\end{equation}
with node set:
\begin{equation}
    \mathcal{V}
    =
    \mathcal{V}_{U}
    \cup
    \mathcal{V}_{T}
    \cup
    \mathcal{V}_{O}
    \cup
    \mathcal{V}_{H},
    \label{eq:graph_nodes}
\end{equation}
where $\mathcal{V}_{U}$ contains UAV nodes, $\mathcal{V}_{T}$ contains 
executable tasks, $\mathcal{V}_{O}$ contains grounded semantic objects or 
reconnaissance elements, and $\mathcal{V}_{H}$ represents home locations.

Edges in $\mathcal{E}$ describe mission relationships including capability 
eligibility, semantic association, precedence, and spatial reachability. A 
parent semantic task may therefore correspond to several executable object 
targets or reconnaissance patches while retaining a common mission identity.

For UAV $u_i$ and task $\tau_j$, capability eligibility is defined as:
\begin{equation}
    e^{\mathrm{cap}}_{ij}
    =
    \begin{cases}
        1,
        & \mathcal{C}^{\mathrm{req}}_j
        \subseteq
        \mathcal{C}_i,
        \\
        0,
        & \mathrm{otherwise}.
    \end{cases}
    \label{eq:capability_eligibility}
\end{equation}

Capability compatibility alone does not establish executability. Spatial and 
energy feasibility are evaluated before the corresponding UAV task relationship 
is admitted to the allocation problem.

\subsection{Physical Feasibility and Energy Aware Planning}
\label{subsec:physical_feasibility}
For each candidate-UAV task pair, AgenticSwarm evaluates obstacle aware 
reachability using the navigation map. Let $\mathbf{g}_j$ denote the spatial 
goal associated with task $\tau_j$. The corresponding path is:
\begin{equation}
    P_{ij}
    =
    A^{*}
    \left(
        \mathbf{p}_i,\,
        \mathbf{g}_j,\,
        \mathcal{M}_{\mathrm{nav}}
    \right).
    \label{eq:astar_path}
\end{equation}

If no valid path exists, the candidate UAV task relation is removed from the 
feasible allocation set. For a valid route, its length:
\begin{equation}
    d_{ij}
    =
    \sum_{k=1}^{|P_{ij}|-1}
    \left\|
        \mathbf{p}_{k+1}-\mathbf{p}_{k}
    \right\|_2
    \label{eq:path_distance}
\end{equation}
is used to estimate travel time and energy requirements.
Return feasibility is evaluated before assignment. A task is energy feasible 
for UAV $u_i$ when:
\begin{equation}
    \hat{E}(P_{ij})
    +
    \hat{E}_{j}
    +
    \hat{E}(P_{jH})
    +
    E^{\mathrm{res}}_i
    \leq
    b_i ,
    \label{eq:rth_constraint}
\end{equation}
where $\hat{E}(P_{ij})$ denotes the estimated travel energy to the task, 
$\hat{E}_{j}$ is the execution cost of the task, $P_{jH}$ represents a feasible 
return route from the task to home, and $E^{\mathrm{res}}_i$ is the protected 
battery reserve.

The current implementation evaluates energy using a normalized simulation model. 
The formulation therefore provides energy aware feasibility rather than a 
physically calibrated battery discharge prediction. The same planning interface 
permits a calibrated vehicle energy model to replace this estimator for physical 
deployment.

Combining capability and physical feasibility gives:
\begin{equation}
    f_{ij}
    =
    e^{\mathrm{cap}}_{ij}
    \,
    e^{\mathrm{path}}_{ij}
    \,
    e^{\mathrm{energy}}_{ij},
    \label{eq:combined_feasibility}
\end{equation}
where $f_{ij}=1$ indicates that UAV $u_i$ is eligible for task $\tau_j$ under 
the current mission state.

\subsection{Constrained Task Allocation and Route Generation}
\label{subsec:task_allocation}
The feasible mission graph is passed to an OR-Tools based optimization stage. 
The optimizer determines which tasks are executed, which UAV performs each 
selected task, and the corresponding execution order.
AgenticSwarm then searches for a mission plan $\Pi$ within the feasible plan set $\mathcal{F}$:
\begin{equation}
    \Pi^{*}
    =
    \arg\min_{\Pi \in \mathcal{F}}
    J(\Pi),
    \label{eq:allocation_opt}
\end{equation}
with a weighted objective of the form:
\begin{equation}
\begin{split}
    J(\Pi)
    ={}&
    w_{m} C_{\mathrm{make}}(\Pi)
    +
    w_{r} C_{\mathrm{route}}(\Pi)
    \\
    &+
    w_{a} C_{\mathrm{reassign}}(\Pi)
    +
    w_{u} C_{\mathrm{unsat}}(\Pi),
\end{split}
    \label{eq:allocation_objective}
\end{equation}
where $C_{\mathrm{make}}$ represents mission duration, 
$C_{\mathrm{route}}$ captures route related cost, 
$C_{\mathrm{reassign}}$ penalizes unnecessary assignment changes during 
replanning, and $C_{\mathrm{unsat}}$ penalizes mission work that remains 
unsatisfied. The corresponding weights determine their relative influence.

The feasible set $\mathcal{F}$ incorporates capability restrictions, task 
precedence, maximum route duration, energy capacity, protected return reserve, 
and task specific coupling constraints. Pickup and delivery operations can, for 
example, require the same UAV to execute both related tasks. Required work is 
assigned a high unsatisfied penalty so that the optimizer favors its completion 
while still permitting a maximum feasible partial mission when the complete task 
set is physically impossible.

The resulting discrete routes are converted into execution trajectories using 
collision safe smoothing. Optional spatial deconfliction can further enforce 
horizontal separation or altitude separation between UAV trajectories before 
the final mission package is produced.

\subsection{Mission Execution and Residual Replanning}
\label{subsec:residual_replanning}
The execution layer receives the ordered UAV routes and maintains the evolving 
mission state. The framework records current UAV position, battery state, 
operational status, completed tasks, completed semantic targets, and 
reconnaissance progress. The execution interface can operate through the 
mathematical mission emulator or through the ROS~2 integration layer connected 
to Gazebo, MAVROS, and ArduPilot.

Runtime events can invalidate part of an active plan. These events include UAV 
failure, battery degradation, vehicle restoration, mission task modification, 
and safety driven return. AgenticSwarm therefore constructs a residual mission 
from the current execution state.

Let $\mathcal{T}$ denote the active mission task set, 
$\mathcal{T}^{C}_{t}$ the completed tasks at time $t$, and 
$\mathcal{T}^{X}_{t}$ the tasks that have been cancelled or removed. The 
remaining task set is:
\begin{equation}
    \mathcal{T}^{R}_{t}
    =
    \mathcal{T}
    \setminus
    \left(
        \mathcal{T}^{C}_{t}
        \cup
        \mathcal{T}^{X}_{t}
    \right).
    \label{eq:residual_tasks}
\end{equation}

Residual reconstruction additionally preserves spatial mission progress. Let 
$\mathcal{P}^{C}_{t}$ represent completed object targets and reconnaissance 
patches. The replanning state is then:
\begin{equation}
    \mathcal{S}_{t}
    =
    \left\{
        \mathcal{T}^{R}_{t},\,
        \mathcal{U}_{t},\,
        \mathcal{P}^{C}_{t},\,
        \mathcal{M}_{\mathrm{nav}}
    \right\},
    \label{eq:residual_state}
\end{equation}
where $\mathcal{U}_{t}$ contains the current state of each UAV.

A residual mission graph $\mathcal{G}_{t}^{R}$ is reconstructed from 
$\mathcal{S}_{t}$. Previously completed semantic targets and reconnaissance 
patches remain satisfied, while feasibility for unfinished work is recomputed 
from the current UAV positions and battery states. The updated graph is then 
passed again through physical feasibility analysis and constrained allocation:
\begin{equation}
    \mathcal{G}_{t}^{R}
    \rightarrow
    \mathcal{F}_{t}^{R}
    \rightarrow
    \Pi_{t}^{*}.
    \label{eq:residual_replan}
\end{equation}

The candidate residual plan is installed only after the reconstructed mission
representation passes validation and the optimizer returns an acceptable plan. Algorithm~\ref{alg:agenticswarm} summarizes the complete AgenticSwarm
planning and adaptation procedure.

\begin{algorithm}[t]
\small
\caption{AgenticSwarm Mission Planning and Adaptation}
\label{alg:agenticswarm}
\begin{algorithmic}[1]

\Require aerial observation $I$, mission instruction $q$, UAV team $\mathcal{U}$
\Ensure executable mission plan $\Pi^{*}$

\State $\mathcal{M}_{\mathrm{sem}} \gets \Psi_{\mathrm{sem}}(I)$
\State $\mathcal{M}_{\mathrm{nav}} \gets \Phi(\mathcal{M}_{\mathrm{sem}})$
\State Interpret $q$ and ground mission targets in $\mathcal{M}_{\mathrm{sem}}$
\State Construct grounded mission graph $\mathcal{G}$

\ForAll{candidate UAV-task pairs $(u_i,\tau_j)$}
    \State Compute $P_{ij} \gets A^{*}
    (\mathbf{p}_i,\mathbf{g}_j,\mathcal{M}_{\mathrm{nav}})$
    \State Evaluate $f_{ij}$ from capability, path, energy, and RTH feasibility
\EndFor

\State $\mathcal{F} \gets \{(i,j)\mid f_{ij}=1\}$
\State $\Pi^{*} \gets \arg\min_{\Pi\in\mathcal{F}} J(\Pi)$
\State Generate routes and execute $\Pi^{*}$

\If{mission or fleet state changes}
    \State Preserve completed tasks and spatial progress $\mathcal{P}^{C}_{t}$
    \State $\mathcal{T}^{R}_{t}
    \gets \mathcal{T}\setminus
    (\mathcal{T}^{C}_{t}\cup\mathcal{T}^{X}_{t})$
    \State Update current UAV state $\mathcal{U}_{t}$
    \State Rebuild residual mission graph $\mathcal{G}^{R}_{t}$
    \State Recompute feasibility set $\mathcal{F}^{R}_{t}$
    \State $\Pi^{*}_{t}
    \gets \arg\min_{\Pi\in\mathcal{F}^{R}_{t}}J_{t}(\Pi)$
    \State Install $\Pi^{*}_{t}$ if the residual plan is valid
\EndIf

\end{algorithmic}
\end{algorithm}

\section{Experimental Setup}
\label{sec:exp-setup}

\subsection{UAV Task Scenarios}
\label{sec:uav-scenarios}
\begin{figure*}[h!]
    \centering
    \includegraphics[width=0.94\textwidth]{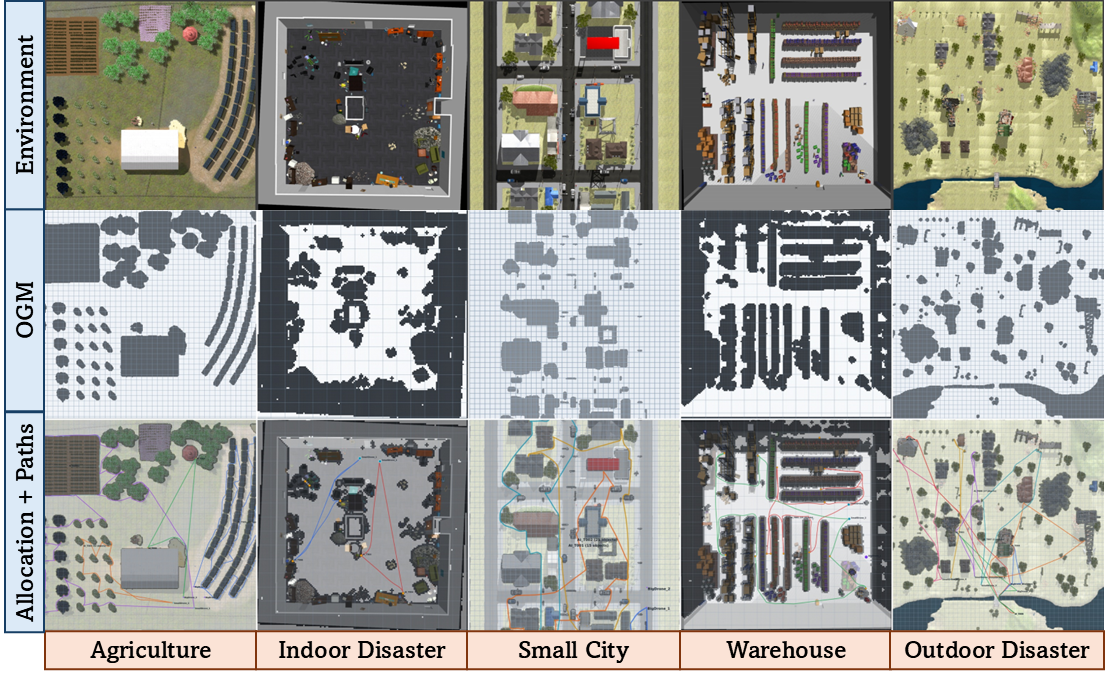}
    \caption{Representative environments used to evaluate heterogeneous
    UAV-swarm missions: (a) agriculture, (b) earthquake-damaged office,
    (c) small city, (d) warehouse, and (e) fire outbreak.}
    \label{fig:uav_scenarios}
    \vspace{-0.4cm}
\end{figure*}

We evaluated AgenticSwarm using Gazebo Classic~11.15.1 across five representative operating environments: an agricultural site, an earthquake-damaged office, a small urban neighborhood, a warehouse, and an outdoor fire outbreak search-and-rescue environment (Fig.~\ref{fig:uav_scenarios}). These environments include both indoor
and outdoor settings with buildings, vegetation, roads, vehicles, people,
debris, storage infrastructure, water, smoke, and fire hazards. Across the
experiments, the mission specifications exercised point target visitation and
inspection, area reconnaissance and search, priority constrained payload
delivery, coupled pickup-and-delivery operations such as water collection and
fire suppression, obstacle and no-fly zone avoidance, and safe return to home.
Runtime events, including UAV failure, battery degradation, and mission
modification, were additionally used to evaluate residual task reallocation.
Together, the scenarios cover a broad cross section of perception, planning,
coordination, logistics, and emergency response tasks relevant to heterogeneous
UAV swarms.

\subsection{Evaluation Metrics}
\label{sec:evaluation-metrics}
Semantic perception was evaluated against 285 manually annotated references
across five environments, comprising 272 obstacle instances and 13 traversable regions. Free form labels from the reference and both perception pipelines
were normalized to a common mission oriented ontology before evaluation.

Predictions were matched one to one with the reference annotations using
Hungarian assignment. A prediction was considered correct when the normalized
semantic class matched the reference and the bounding box intersection over
union (IoU) was at least 0.5. Precision, recall, and F1 were computed as:
\begin{equation}
\small
P=\frac{TP}{TP+FP},\;
R=\frac{TP}{TP+FN},\;
F_1=\frac{2PR}{P+R},
\label{eq:prf1}
\end{equation}
where $TP$, $FP$, and $FN$ denote matched predictions, unmatched predictions,
and unmatched reference instances, respectively.

In addition to class aware detection performance, semantic label accuracy was
computed over successfully localized matches. Mission critical recall was
evaluated over people, vehicles, damaged structures, debris, fire, and smoke,
since these classes directly influence downstream mission grounding.

The navigation representation was evaluated using occupancy envelope IoU and
false free rate. The latter measures reference obstacle space that is
incorrectly represented as traversable. All reported values are macro averaged
across the five environments.

\section{Experimental Results}
\label{sec:result}

\subsection{Semantic Perception Results}
\label{sec:semantic-results}

To evaluate the choice of the semantic perception backbone, the final
SAM3 based pipeline was compared against an alternative perception stack
combining Grounding DINO for open-vocabulary object detection with SAM~2.1
for segmentation~\cite{liu2024grounding,ravi2025sam2}. Both pipelines were
evaluated using the same annotated environments and metrics.

Table~\ref{tab:semantic_accuracy} shows that the SAM3 based pipeline achieved
higher performance across all aggregated semantic metrics. Class-aware recall
increased by $25.2$ percentage points (pp), from $21.0\%$ to $46.2\%$, while
F1 increased by $11.7$~pp. Semantic label accuracy increased by $29.5$~pp, and
recall over mission-critical classes improved by $24.3$~pp. These results
motivated the use of SAM3 as the primary semantic perception component in the
final framework.
\begin{table}[h!]
\centering
\footnotesize
\setlength{\tabcolsep}{4pt}
\caption{Semantic perception performance averaged across five
environments. $\Delta$ is the absolute change in percentage points.}
\label{tab:semantic_accuracy}
\begin{tabular}{lrrr}
\hline
Metric & GDINO+SAM2.1 & SAM3 & $\Delta$ \\
\hline
Precision (\%)        & 12.5 & \textbf{20.2} & +7.7  \\
Recall (\%)           & 21.0 & \textbf{46.2} & +25.2 \\
F1 (\%)               & 15.4 & \textbf{27.1} & +11.7 \\
Label accuracy (\%)   & 66.1 & \textbf{95.6} & +29.5 \\
Critical recall (\%)  & 10.2 & \textbf{34.5} & +24.3 \\
\hline
\end{tabular}
\end{table}

Performance gains varied across environments. F1 increased by $35.0$~pp in the
agriculture scene and by $25.6$~pp in the small city scene, while warehouse
performance remained nearly unchanged. The earthquake office was the main
failure case, where F1 decreased by $10.6$~pp. This degradation was primarily
associated with debris fragmentation, which generated 165 predicted obstacle
instances for 21 annotated objects. Overall, SAM3 produced 698 obstacle
instances compared with 423 from the Grounding DINO+SAM~2.1 baseline,
indicating that the increased recall is accompanied by remaining fragmentation
and spurious detections.

The improved semantic perception also produced a stronger navigation
representation. Occupancy envelope IoU increased from $36.8\%$ to $55.8\%$
($+19.0$~pp), while the false-free rate decreased from $53.3\%$ to $29.5\%$
($-23.8$~pp). Both metrics improved across all five environments. The reduction
in false free space is particularly important for downstream path planning,
where missed obstacle regions can lead to infeasible route estimates.

\subsection{Agentic Task Allocation and Mission Planning}
\label{sec:agentic_allocation_results}
One representative natural language mission was evaluated in each environment.
Together, the missions exercised capability aware and dependency constrained
allocation, multi object tasks, obstacle aware planning, UAV failure, battery
degradation, and online task insertion. Table~\ref{tab:mission_case_studies}
summarizes the grounded objectives and the corresponding agentic response.
\begin{table*}[h!]
\centering
\scriptsize
\setlength{\tabcolsep}{3pt}
\renewcommand{\arraystretch}{1.12}
\caption{Qualitative summary of the five representative mission
executions. ``Affected tasks'' denotes unfinished work assigned to a UAV when
a failure or battery event occurred.}
\label{tab:mission_case_studies}
\begin{tabularx}{\textwidth}{@{}lXXX@{}}
\toprule
Environment & Grounded mission & Runtime change & AgenticSwarm response \\
\midrule
Agriculture &
Clean reachable solar-panel rows, scan the crop field and vegetation, perform
an ordered payload transfer, and return home. &
BigDrone\_3 battery reduced to 20.5\%; SmallDrone\_2 failed. &
Sent the battery-degraded UAV home, preserved completed work, and reassigned
all 11 affected tasks to eligible UAVs. \\

Earthquake office &
Visit six independently grounded people using two camera-and-indicator UAVs
and return home. &
No runtime event. &
Completed all four reachable person visits and explicitly deferred two targets
for which no collision-free route existed. \\

Warehouse &
Inspect every reachable storage rack with inspection UAVs and return home. &
A box pickup-and-delivery task was inserted at $t=8$~s. &
Assigned the inserted payload task to the capable heavy UAV and completed it;
one unreachable rack target was deferred. \\

Small city &
Acquire number-plate imagery for 15 vehicles, roof imagery for 21 structures,
and inspect two bridges. &
SmallDrone\_4 and SmallDrone\_5 failed at $t=10$~s and $t=20$~s. &
Preserved ongoing work and reassigned all 16 affected tasks; two structure
targets with no collision-free route remained deferred. \\

Fire Outbreak &
Collect water before suppressing three fires, inspect two damaged structures,
deliver medical supplies to all grounded people, avoid smoke, and return home. &
SmallDrone\_5 failed; SmallDrone\_3 battery fell to 21\%; an SOS task was
inserted. &
Returned the battery-degraded UAV, reassigned all nine affected tasks, and
inserted and completed the new SOS delivery task. \\
\bottomrule
\end{tabularx}
\end{table*}

Table~\ref{tab:mission_quantitative} reports executable leaf task outcomes;
aggregate parent tasks are excluded to prevent double counting, and failed UAVs
are excluded from the return home denominator.
\begin{table}[h!]
\centering
\scriptsize
\setlength{\tabcolsep}{2pt}
\renewcommand{\arraystretch}{1.08}
\caption{Quantitative outcomes for the representative recorded runs.
Affected-task recovery excludes newly inserted tasks, which are reported
separately.}
\label{tab:mission_quantitative}
\begin{tabularx}{\columnwidth}
{@{}>{\raggedright\arraybackslash}p{0.34\columnwidth}YYYYY@{}}
\toprule
Metric & Agriculture & Office & Warehouse & City & Fire \\
\midrule
UAVs                     & 5     & 2    & 3     & 8     & 8     \\
Tasks completed          & 95/95 & 4/6  & 15/16 & 36/38 & 36/36 \\
Runtime events           & 2     & 0    & 1     & 2     & 3     \\
Affected tasks recovered & 11/11 & --   & --    & 16/16 & 9/9   \\
Inserted tasks completed & --    & --   & 1/1   & --    & 1/1   \\
Mission time (s)         & 245.6 & 21.0 & 60.0  & 157.0 & 206.2 \\
Operational UAVs home    & 4/4   & 2/2  & 3/3   & 6/6   & 7/7   \\
\bottomrule
\end{tabularx}
\end{table}

Across the three scenarios with UAV-state disruptions, all 36 affected tasks
were recovered, both inserted tasks were completed, and all 22 operational UAVs
returned home. The five uncompleted tasks were unreachable targets rather than
work abandoned following a runtime event.

Table~\ref{tab:inference_overhead} reports language model usage for perception
and residual replanning. Replanning tokens are cumulative within each mission,
whereas latency is the mean per runtime event; initial mission interpretation is
excluded because previously prepared plans were used in four environments.
\begin{table}[h!]
\centering
\scriptsize
\setlength{\tabcolsep}{2pt}
\renewcommand{\arraystretch}{1.08}
\caption{Recorded language-model inference overhead. Perception
latency is cumulative over two calls; replanning tokens are cumulative over
the listed events, and replan latency is the mean per event.}
\label{tab:inference_overhead}
\begin{tabularx}{\columnwidth}
{@{}>{\raggedright\arraybackslash}p{0.34\columnwidth}YYYYY@{}}
\toprule
Metric & Agriculture & Office & Warehouse & City & Fire \\
\midrule
Perception tokens       & 16,299 & 18,965 & 13,968 & 15,811 & 36,082 \\
Perception latency (s)  & 117.1  & 136.4  & 89.2   & 102.9  & 172.5  \\
Replan calls            & 2      & 0      & 1      & 2      & 3      \\
Replanning tokens       & 22,482 & --     & 3,943  & 16,205 & 23,432 \\
Mean replan latency (s) & 20.6   & --     & 10.8   & 24.9   & 19.5   \\
\bottomrule
\end{tabularx}
\end{table}

The wildfire scene incurred the largest perception workload. Across missions
with runtime events, mean replanning latency ranged from 10.8~s to 24.9~s.

\subsection{Real-World Implementation}
\label{sec:real-world-implementation}
\begin{figure*}[h!]
    \centering
    \includegraphics[width=0.98\textwidth]{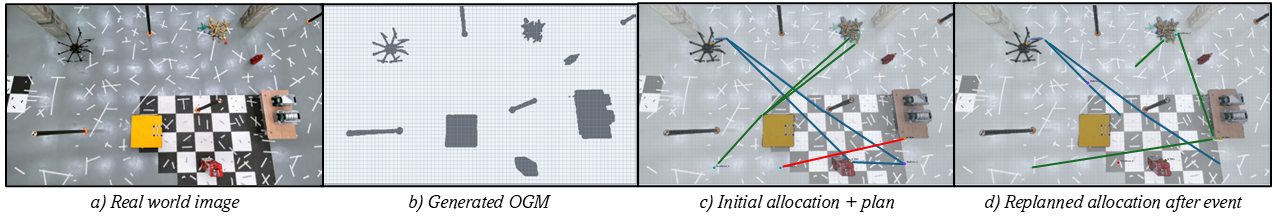}
    \caption{Semantic perception of the real world scenario with the resulting task allocation and planned paths.}
    \label{fig:real_world_plan}
    \vspace{-0.4cm}
\end{figure*}

The framework was further demonstrated in an indoor flight arena using a
top down RGB observation of the physical scene. The perception pipeline
identified the multirotor, red payload boxes, scattered debris, wooden table,
and vertical pillars, and converted these detections into a navigation OGM.
The natural language mission requested an ordered battery pickup and delivery,
debris inspection with a scanner equipped UAV, standoff photography of the
table contents, avoidance of the pillars, and return to home. AgenticSwarm
grounded these objectives and generated capable and path feasible routes
for one delivery UAV and two sensing UAVs as illustrated in Fig.~\ref{fig:real_world_plan}.

\begin{figure*}[h!]
    \centering
    \includegraphics[width=0.98\textwidth]{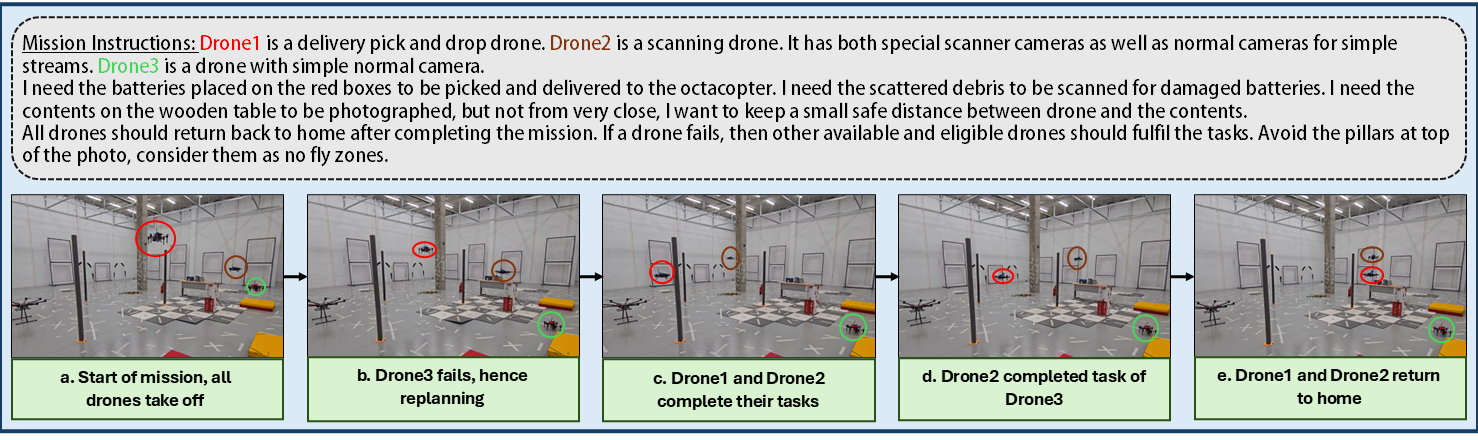}
    \caption{Real world mission execution from the natural language instruction.}
    \label{fig:real_exp}
    \vspace{-0.4cm}
\end{figure*}

Fig.~\ref{fig:real_exp} shows the resulting execution sequence. A failure was
triggered for the UAV initially assigned to the photography task after takeoff,
while its task was unfinished. Residual replanning removed the failed vehicle, retained
the active delivery and scanning progress, and reassigned the photography task
to the remaining UAV carrying a conventional camera. The delivery UAV
continued its pickup--delivery route, the replacement UAV completed the
reassigned observation, and both operational UAVs returned home. This
trial demonstrates transfer from perception of a physical scene to
capability aware mission execution and failure recovery, while also showing
that the framework reports spatially infeasible work rather than forcing an
unsafe assignment.

\subsection{Ablation Study}
\label{sec:ablation-results}

To isolate the contribution of the main framework components, three ablated
variants of AgenticSwarm were evaluated, each removing one component while
retaining the same frozen OGM, mission specification, fleet state, and event
condition. Table~\ref{tab:ablation} reports one execution per environment.
Recovery statistics are aggregated over the three environments with valid paired comparisons (agriculture, warehouse,
and small city).

\begin{table}[h!]
\centering
\scriptsize
\setlength{\tabcolsep}{3pt}
\renewcommand{\arraystretch}{1.12}
\caption{Component ablation study of AgenticSwarm. Each ablated variant removes
one component from the full framework while all other experimental conditions
are held fixed.}
\label{tab:ablation}
\begin{tabularx}{\columnwidth}{@{}p{0.30\columnwidth}X@{}}
\toprule
Ablated variant & Effect relative to the full system \\
\midrule

w/o capability \& dependency reasoning &
Semantically valid allocations decreased from 5/5 to 1/5 environments;
74 capability mismatches, 1 dependency violation, and 4 coupling violations
were introduced. \\

w/o path-aware allocation &
Executable route sets decreased from 5/5 to 2/5 environments. For the two
executable pairs, distance increased by 42.3\% (agriculture) and 3.6\%
(fire outbreak). \\

w/o residual replanning &
Mean repeated work increased from 0\% to 61.7\%; pooled travel distance and
post-event recovery time increased by 41.9\% and 58.6\%, respectively
($n=3$ paired environments). \\

\bottomrule
\end{tabularx}
\end{table}

The ablations reveal complementary roles of the three components. Capability
and dependency reasoning prevents invalid UAV--task assignments, path-aware
allocation accounts for route feasibility during assignment, and residual
replanning limits repeated work and additional recovery cost following runtime
events.


\section{Limitations}
\label{sec:limitations}
AgenticSwarm currently relies on a predominantly static 2D OGM, assumed vertical clearance, and a simplified energy model; deployment in new environments therefore requires spatial calibration and vehicle specific validation. Performance also depends on semantic grounding accuracy, while cloud inference introduces delays during replanning. Finally, evaluation used one representative mission per environment and a limited physical demonstration, motivating repeated real world trials with larger fleets and more varied operating conditions.


\section{Conclusion \& Future Work}
This paper presented AgenticSwarm, a framework that connects open vocabulary
semantic perception and natural language mission grounding with explicit
capability, dependency, path, energy, and return home constraints for
heterogeneous UAV teams. Unlike approaches that use language output directly
as a flight plan, AgenticSwarm restricts the agent to semantic mission
construction and validates assignments through A*, constrained optimization,
and execution state aware residual replanning.

Across five heterogeneous environments, the SAM3 perception pipeline improved
class aware recall by $25.2$~pp and semantic label accuracy by $29.5$~pp over the
Grounding DINO+SAM~2.1 baseline. Runtime recovery reassigned all 36 affected
tasks, completed both inserted tasks, and returned all 22 operational UAVs
home. The ablations further showed that removing semantic constraints reduced
valid allocations from 5/5 to 1/5 environments, removing path awareness
reduced executable route sets from 5/5 to 2/5, and removing residual replanning
increased repeated work to $61.7\%$ and post-event recovery time by $58.6\%$.
The indoor demonstration additionally confirmed capability aware reassignment
after a UAV failure in a scene derived from real imagery.

Future work will incorporate online semantic-map updates and calibrated
image-to-world localization. Onboard or locally hosted inference and
incremental optimization will be investigated to reduce recovery latency,
followed by repeated hardware trials with larger fleets and dynamic obstacles.

\balance

\bibliographystyle{IEEEtran}
\bibliography{IEEEexample} 
\end{document}